\documentclass[10pt,twocolumn,letterpaper]{article}

\usepackage[pagenumbers]{cvpr} 

\definecolor{cvprblue}{rgb}{0.21,0.49,0.74}
\usepackage[pagebackref,breaklinks,colorlinks,allcolors=cvprblue]{hyperref}

\usepackage{amsmath}

\usepackage{amssymb}
\usepackage{mathtools}
\usepackage{amsthm}
\usepackage{pifont}
\usepackage{textcomp}
\usepackage{caption}
\usepackage[table,xcdraw]{xcolor}
\usepackage{multirow,booktabs,makecell}
\usepackage{setspace}
\usepackage{threeparttable}
\usepackage{float}
\usepackage{stfloats}
\usepackage{gensymb}
\usepackage{comment}
\usepackage{algorithm}
\usepackage{algorithmicx}
\usepackage{algpseudocode}
\usepackage{arydshln}
\usepackage{url}
\usepackage{soul}   
\soulregister\cite7
\soulregister\ref7
\usepackage{tabularx}
\definecolor{Gray}{gray}{0.5}   
\usepackage{colortbl}
\usepackage{adjustbox}
\usepackage{enumitem}
\usepackage{pifont}
\usepackage{upquote}  
\usepackage{hhline}
\definecolor{myblue}{RGB}{0,100,200}
\definecolor{mygreen}{HTML}{009900}
\definecolor{mygray}{gray}{0.5}
\definecolor{mypurple}{HTML}{CDA0CD}

\definecolor{BestColor}{HTML}{ffffff}
\definecolor{SecondColor}{HTML}{ffffff}

\usepackage{bm}

\usepackage{hyperref}
\usepackage{url}
\usepackage{wrapfig}
\usepackage{placeins}

\def\confName{CVPR}
\def\confYear{2027}

\title{\textsc{WorldWeave}: Rethinking Embodied Navigation through \\ Cross-Task World Modeling}

\author{
    Xuan Yao\textsuperscript{1,2,3}\thanks{Equal contribution.} \quad
    Jiahao Zhang\textsuperscript{1,3$*$} \quad
    Rongjunchen Zhang\textsuperscript{3$*$} \\
    Kaixuan Huang\textsuperscript{3} \quad
    Junyu Gao\textsuperscript{1,2}\thanks{Corresponding author.} \quad
    Changsheng Xu\textsuperscript{1,2,4} \\
    \textsuperscript{1}State Key Laboratory of Multimodal Artificial Intelligence Systems (MAIS), \\ Institute of Automation, Chinese Academy of Sciences, Beijing, China\\
    \textsuperscript{2}School of Artificial Intelligence, University of Chinese Academy of Sciences (UCAS), Beijing, China\\
    \textsuperscript{3}HiThink Research, China \qquad
    \textsuperscript{4}Peng Cheng Laboratory, Shenzhen, China\\
    {\tt\small yaoxuan2022@ia.ac.cn, jahaozhang@163.com, \{zhangrongjunchen, huangkaixuan\}@myhexin.com} \\
    {\tt\small junyu.gao@nlpr.ia.ac.cn, csxu@nlpr.ia.ac.cn}
}

\begin{document}
\maketitle
\begin{abstract}
Mobile embodied agents continuously alter their observations through their own motion, while external dynamics further shape future interactions. Different mobile embodied tasks therefore capture complementary patterns of state evolution, yet jointly learning world dynamics across such interactions remains less explored. We investigate this problem through Vision-and-Language Navigation (VLN) and Embodied Visual Tracking (EVT), which respectively involve navigation toward a stationary goal and interaction with a moving target. We propose \textup{\textsc{WorldWeave}}, a cross-task world model that {couples shared predictive dynamics with task-specific adaptation.}
Instead of conditioning state transitions on heterogeneous physical actions, \textup{\textsc{WorldWeave}} learns latent actions from observed state changes via inverse-forward dynamics, {establishing a shared transition representation across tasks. The resulting future-state predictions are adaptively integrated into task-specific representations for downstream decision making. Experiments across multiple VLN and EVT benchmarks demonstrate consistent improvements over specialized baselines.}
\end{abstract}

\section{Introduction}\label{sec:intro}

\begin{figure}[t]
    \centering
    \includegraphics[width=\linewidth]{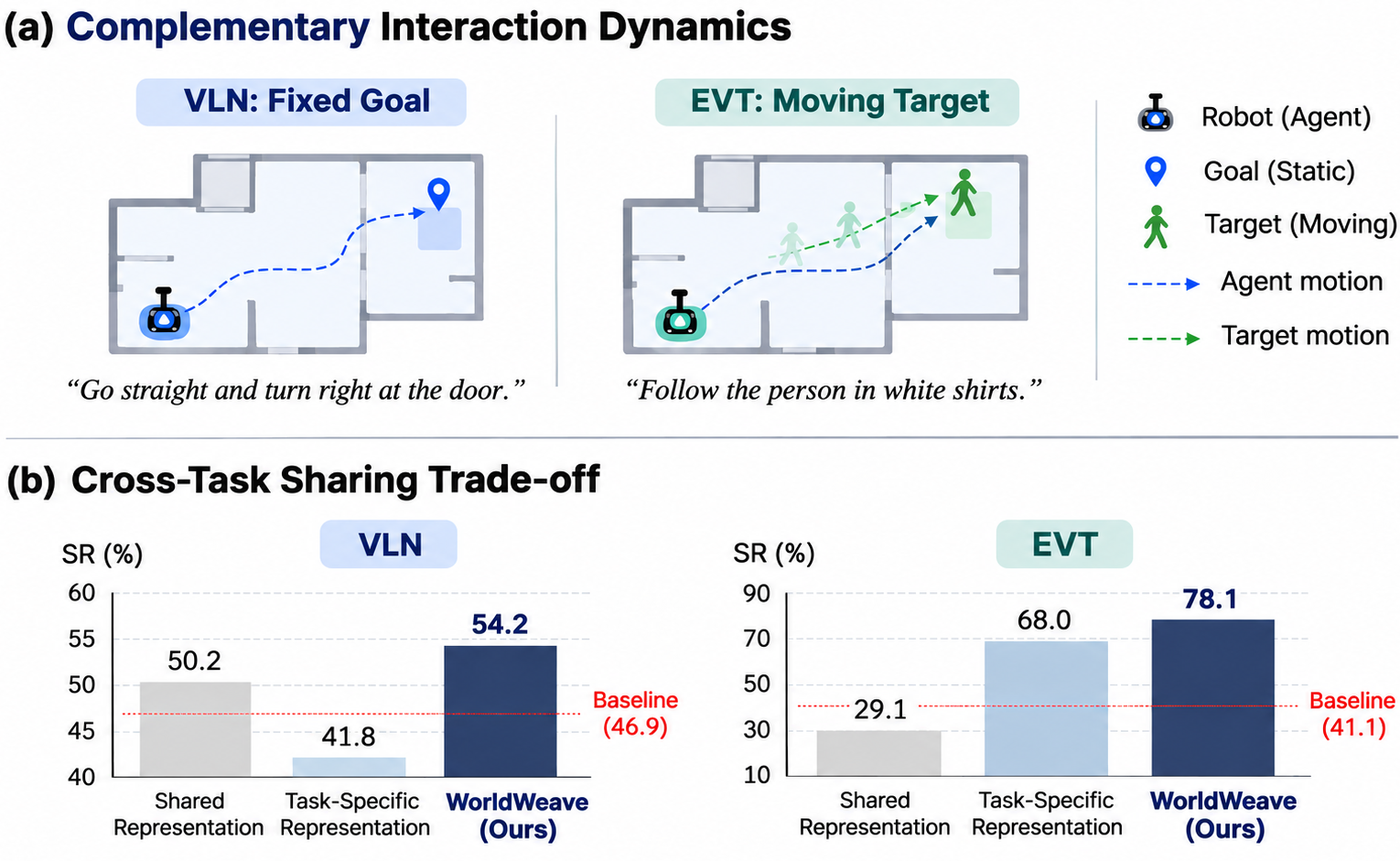}
    \vspace{-0.5cm}
\caption{
\textbf{Motivation of \textup{\textsc{WorldWeave}}.}
(a) VLN and EVT capture complementary dynamics through navigation to fixed goals and tracking moving targets.
(b) Naive representation sharing creates cross-task trade-offs, which \textup{\textsc{WorldWeave}} mitigates through shared predictive dynamics and task-specific adaptation.
}
    \vspace{-0.5cm}
    \label{fig:intro}
\end{figure}

A mobile robot's observations are inherently shaped by its own actions: each movement changes what the agent perceives and alters its spatial relationship with the environment and task goal. This creates a continuous perception--action loop, in which current decisions shape future observations, which in turn inform subsequent decisions. Effective mobile decision making therefore requires not only understanding the current interaction state, but also anticipating how that state may evolve as the agent acts.
The dynamics underlying this evolution can vary substantially with the task goal and interaction setting. While an agent's own motion continuously changes its visual observations and spatial relationships with the environment, a moving target introduces an additional source of state variation. Vision-and-Language Navigation (VLN)~\citep{Visionandlanguagenavigation,Beyondthenavgraph} and Embodied Visual Tracking (EVT)~\citep{zhong2019ad,trackvla} represent two complementary settings that capture these different aspects of mobile embodied interaction. In VLN, an agent grounds natural-language instructions in visual observations and navigates toward an instruction-specified goal that remains fixed within an episode. Its state evolution is primarily driven by the agent's own motion as it progresses through environmental structures over extended trajectories. EVT, in contrast, requires a mobile agent to actively follow a moving target through continuous perception and control, where the evolving interaction state is jointly shaped by the agent's movement and the target's independent motion.

Although VLN and EVT differ in task objectives and interaction patterns, both involve mobile agent-environment interaction. In VLN, state evolution primarily reflects the agent's progression through environmental structures over extended trajectories~\citep{navmorph,zhang2026futurenav}, whereas in EVT, it additionally involves the continuously changing agent--target relationship shaped by both agent and target motion~\citep{uss}. As illustrated in Fig.~\ref{fig:intro}\,(a), the two tasks offer complementary perspectives on embodied state evolution. 
Together, they provide an opportunity to study world modeling beyond the dynamics of either task alone. Despite their distinct goals and action spaces, can these related interaction patterns be leveraged to learn shared predictive dynamics?

A natural framework for modeling such predictive dynamics is a world model~\citep{worldmodels}, which learns how latent states evolve through interaction.
Conventional world models typically model state transitions conditioned on predefined actions, \textit{i.e.}, $p(\boldsymbol{s}_{t+1}\,|\,\boldsymbol{s}_t,\boldsymbol{a}_t)$~\citep{learning,dream}.
While well suited to individual tasks, extending it across heterogeneous tasks introduces additional challenges, as the actions conditioning state transitions may differ across task settings~\citep{zhang2026mixture}.
In our setting, directly conditioning a shared world model on the heterogeneous actions of VLN and EVT would tie its learned dynamics to task-specific action definitions.
Latent-action learning provides an alternative by inferring latent transition variables from state changes encoded in consecutive observations, rather than relying on predefined action labels~\citep{genie,gao2026dreamdojo}. This motivates a shared transition representation across heterogeneous embodied tasks.

Establishing such a transition representation, however, does not determine how the shared dynamics should support each task. VLN and EVT rely on predictive dynamics differently for downstream decision making, while naive sharing across heterogeneous tasks can introduce interference and negative transfer~\citep{jeong2024quantifying,jeong2025selective}. Our preliminary experiments make this issue explicit. As shown in {Fig.~\ref{fig:intro}\,(b)}, using a single shared adapter preserves VLN performance but yields little effective tracking capability for EVT. Introducing task-specific adapters substantially recovers EVT performance, whereas directly incorporating the shared world representation into VLN's decision pathway degrades navigation performance. These observations suggest that effective dynamics sharing requires not only a common transition representation, but also {task-specific adaptation of the shared predictions to the distinct representations and decision processes of each task.}

Motivated by these observations, we propose \textsc{WorldWeave}, a cross-task world-modeling framework that {couples shared predictive dynamics with task-specific adaptation for VLN and EVT.}
During pretraining, \textsc{WorldWeave} learns state transitions through inverse--forward dynamics: latent actions are inferred from state changes encoded in consecutive observations and used to predict subsequent states.
Because these latent actions characterize observed state transitions rather than task-specific physical actions, the learned dynamics are not tied to either task's action definition and can be pretrained from videos without action annotations.
To enable prediction without access to future observations at inference time, the model further learns to infer these latent actions from current interaction context.
{The resulting future-state predictions are then adaptively integrated into each task's representation, allowing shared predictive dynamics to support distinct downstream decision processes. In this way, \textsc{WorldWeave} shares how state transitions are modeled while preserving how predictive information is used for task-specific decision making.}

Our contributions are summarized as follows:

\begin{itemize}
    \item We investigate cross-task dynamics sharing in embodied navigation, focusing on Vision-and-Language Navigation (VLN) and Embodied Visual Tracking (EVT) as two complementary interaction settings with distinct forms of state evolution and decision making.
    \item {We propose \textsc{WorldWeave}, which couples shared latent dynamics with task-specific predictive adaptation. It learns latent transitions from state changes in consecutive observations without relying on task-specific action definitions, while adapting shared future-state predictions to distinct task representations and decision processes.}
    \item Extensive experiments across VLN and EVT benchmarks show consistent improvements over task-specific baselines, supporting the effectiveness of cross-task dynamics learning across complementary embodied interactions.

\end{itemize}

\section{Related Work}\label{sec:rw}

\noindent\textbf{Vision-and-Language Navigation.}
Vision-and-Language Navigation (VLN) requires agents to follow natural-language instructions through visual environments, evolving from discrete navigation graphs to continuous environments~\citep{Visionandlanguagenavigation,Beyondthenavgraph}. Subsequent methods incorporate spatial representations and observation history for navigation reasoning~\citep{Thinkglobalactlocal,Gridmm,volumetric,history}, while recent LLM/VLM-based approaches improve instruction understanding and action prediction~\citep{navgpt,navgpt2,mapgpt,abotn0}. Predictive modeling has also been explored, e.g., NavMorph~\citep{navmorph} learns latent state transitions to anticipate future navigation states.

\noindent\textbf{Embodied Visual Tracking.}
Embodied Visual Tracking (EVT) extends active visual tracking to embodied agents that actively follow moving targets through egocentric perception and closed-loop control~\citep{zhong2019ad,zhong2019ad+,trackvla}. Recent methods leverage vision--language--action models for target understanding and motion decision making~\citep{end,trackvla,trackvla++}, while USS~\citep{uss} introduces latent predictive dynamics for temporal modeling. Despite predictive modeling in both VLN and EVT, existing approaches model their dynamics separately, leaving cross-task dynamics sharing largely unexplored.


\noindent\textbf{World Models.}
World models learn predictive state dynamics for sequential decision making, traditionally through action-conditioned latent transitions~\citep{worldmodels,learning,dream}. Recent latent-action approaches instead infer transition variables from temporal visual changes, enabling dynamics learning without predefined action labels~\citep{genie,latent,clam,gao2026dreamdojo}. World models have also been applied to embodied tasks such as navigation and tracking~\citep{astranav,uss}. 
Beyond individual tasks, recent work has explored multi-task world modeling through modular latent dynamics, as exemplified by Mixture-of-World Models~\citep{zhang2026mixture}, which combines task-conditioned dynamics experts with a shared backbone. While this approach organizes multi-task dynamics through modular specialization, \textup{\textsc{WorldWeave}} focuses on sharing predictive dynamics across complementary embodied interactions, while retaining task-specific state representations and predictive adaptation to accommodate distinct decision processes.


\noindent\textbf{Unified Embodied Models.}
Recent generalist embodied models aim to accommodate diverse tasks and embodiments within shared architectures or control interfaces~\citep{octo,scalingcrossembodied,openx,intelligence2025pi,bjorck2025gr00t,guo2026phaselora}. In navigation, NavFoM jointly addresses multiple navigation and tracking tasks, while Qwen-RobotNav unifies navigation settings through a shared waypoint-prediction interface~\citep{embodied,qwenrobotnav}.
 These efforts explore cross-task generalization through shared models or task interfaces. In contrast, {\textup{\textsc{WorldWeave}}} investigates the sharing of predictive dynamics across complementary mobile embodied tasks, while retaining task-specific adaptation to their distinct temporal and decision requirements.

\section{Our Approach}\label{sec:method}

We present \textup{\textsc{WorldWeave}}, a framework that jointly learns predictive world dynamics across Vision-and-Language Navigation (VLN) and Embodied Visual Tracking (EVT) while preserving their task-specific state representations and decision processes.

\noindent\textbf{Problem Formulation.}
We index the two embodied navigation tasks by $\tau\in\{\mathrm{VLN},\mathrm{EVT}\}$ and formulate each task as a partially observable Markov decision process (POMDP) $\mathcal{M}^{\tau}=(\mathcal{S}^{\tau},\mathcal{A}^{\tau},\mathcal{O}^{\tau},\mathcal{T}^{\tau},\Omega^{\tau})$, where $\mathcal{S}^{\tau}$, $\mathcal{A}^{\tau}$, and $\mathcal{O}^{\tau}$ denote the state, action, and observation spaces, respectively, $\mathcal{T}^{\tau}$ and $\Omega^{\tau}$ denote the transition and observation processes.
At time step $t$, the agent receives an observation $\boldsymbol{o}_t^{\tau}\in\mathcal{O}^{\tau}$ and makes a decision based on its observation history $\boldsymbol{o}_{\leq t}^{\tau}$ and natural-language instruction $\boldsymbol{l}^{\tau}$. The instruction provides navigation guidance in VLN and specifies the target to be followed in EVT.
The corresponding policy is denoted by $\pi^{\tau}(\boldsymbol{a}_t^{\tau}\mid\boldsymbol{o}_{\leq t}^{\tau},\boldsymbol{l}^{\tau})$, with $\boldsymbol{a}_t^{\tau}\in\mathcal{A}^{\tau}$.

A shared vision-language backbone \citep{bai2025qwen3} encodes the observation history and instruction into contextual representations $\bm{h}_t^\tau = \text{VLM}(\bm{o}_{\leq t}^\tau, l^\tau)$. We derive the world state $\bm{s}_t^\tau = Q_\tau(\bm{h}_t^\tau)$ for dynamics modeling, where $Q_\tau$ employs learnable world-state queries to extract task-relevant state information from $\bm{h}_t^\tau$ via cross-attention. We denote $\bm{h}_t^\tau \in \mathbb{R}^{N_h^\tau \times d}$ and $\bm{s}_t^\tau \in \mathbb{R}^{N_s^\tau \times d}$. Although state representations may differ in number and semantics across tasks, they lie in a common feature space, allowing state evolution to be modeled through shared predictive dynamics.


\noindent\textbf{Framework Overview.}
VLN and EVT exhibit complementary patterns of state evolution, but their different action semantics and parameterizations make conventional action-conditioned dynamics
$p(\boldsymbol{s}_{t+1}^{\tau}\mid\boldsymbol{s}_t^{\tau},\boldsymbol{a}_t^{\tau})$~\citep{worldmodels,learning,dream}
difficult to share directly across tasks.
Rather than aligning their heterogeneous action spaces, \textup{\textsc{WorldWeave}} represents state changes through a latent action $\boldsymbol{z}_t$, providing a shared transition space for modeling state evolution across tasks without directly conditioning on task-specific physical actions.

\begin{figure*}[t]
    \centering
    \includegraphics[width=\textwidth]{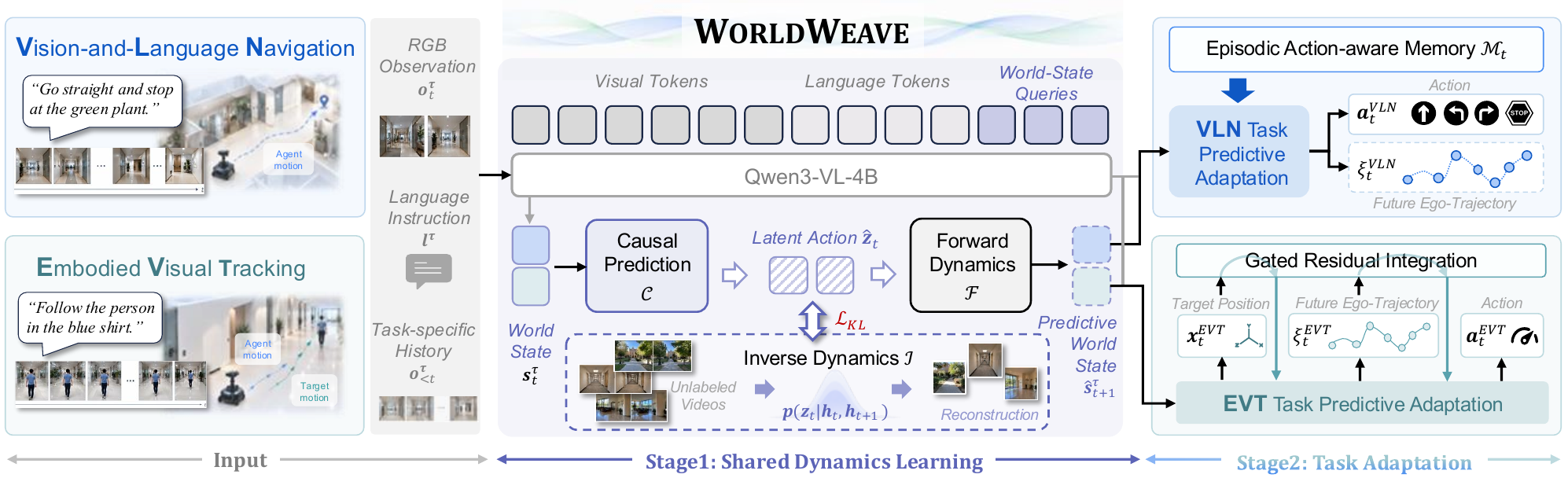}
    \vspace{-0.5cm}
    \caption{
    \textbf{Overview of \textup{\textsc{WorldWeave}}.}
    Shared predictive dynamics are learned from state changes in consecutive
    observations through inverse--forward dynamics and transferred across VLN
    and EVT without relying on task-specific action definitions.
    Task-specific adaptation then incorporates the predicted future states into
    each task's decision process.
    }
    \vspace{0cm}
    \label{fig:method}
\end{figure*}

As illustrated in {Fig.~\ref{fig:method}}, we first learn the shared dynamics from diverse temporally ordered videos without requiring task-specific action annotations.
{Despite their heterogeneous motions, these videos share a common learning signal: how the represented world state changes over time. \textup{\textsc{WorldWeave}} encodes these changes into a shared latent transition space, allowing heterogeneous video transitions to jointly supervise the same transition model.}
Given consecutive visual representations, the model infers a latent action from their temporal change and learns how the represented state evolves under this latent action.
Since future observations are unavailable during online decision making, we further learn a causal latent-action predictor that estimates $\boldsymbol{z}_t$ using only information available up to the current step.
The resulting predictive dynamics can therefore anticipate future states without accessing future observations at decision time.
We detail the learning of these shared dynamics in Sec.~\ref{sec:shared_dynamics}.
The predicted future state is then incorporated differently into each task.
For VLN, \textup{\textsc{WorldWeave}} combines the predicted state with historical context to support navigation, whereas for EVT, it supports structured estimation of target position, ego-trajectory, and continuous control.
These task-specific integration mechanisms are described in Sec.~\ref{sec:task_integration}, followed by the learning objectives and training procedure in Sec.~\ref{sec:training}.

\subsection{Learning Shared World Dynamics}
\label{sec:shared_dynamics}
Building on the formulation above, we use $\boldsymbol{h}_t^\tau$ and $\boldsymbol{s}_t^\tau$ to denote the contextual visual representation and task-specific world-state representation at time $t$, respectively. 
{While latent actions have been used to learn dynamics from videos without action annotations, we leverage them here to enable dynamics sharing across tasks with heterogeneous action spaces. Rather than requiring a unified state or action representation, \textup{\textsc{WorldWeave}} retains task-specific world states while sharing their transition modeling.}
To this end, we introduce a shared latent transition space $\mathcal{Z}$ and formulate predictive state evolution:
\begin{equation}
p(\boldsymbol{s}_{t+1}^{\tau}
\mid
\boldsymbol{s}_t^{\tau},
\boldsymbol{h}_{\leq t}^{\tau})
=
\int_{\mathcal{Z}}
p
\left(
\boldsymbol{s}_{t+1}^{\tau}
\mid
\boldsymbol{s}_t^{\tau},
\boldsymbol{z}_t
\right)
p
\left(
\boldsymbol{z}_t
\mid
\boldsymbol{h}_{\leq t}^{\tau},
\boldsymbol{s}_t^{\tau}
\right)
\,d\boldsymbol{z}_t .
\label{eq:latent_transition_factorization}
\end{equation}
Here, $\boldsymbol{z}_t \in \mathcal{Z}$ serves as a shared latent action that characterizes state changes without relying on task-specific physical action definitions. The transition model $\mathcal{F}$ is shared across tasks, while $\boldsymbol{s}_t^\tau$ retains task-specific state information; the causal predictor $\mathcal{C}$ estimates the latent action from the interaction history available up to the current step. Since the following dynamics-learning formulation applies identically to both tasks, we omit the task superscript $\tau$ for clarity.

\noindent\textbf{Learning Future-Informed Latent Actions.}
During dynamics learning, consecutive observations provide direct evidence of the transition that occurred. We leverage this information through an inverse dynamics model $\mathcal{I}$, which takes the contextual representations before and after the transition to parameterize a future-informed posterior over the latent action:
$\boldsymbol{z}_t
\sim
q
\left(\boldsymbol{z}_t \,|\,
\boldsymbol{h}_t,
\boldsymbol{h}_{t+1}
\right),
\boldsymbol{z}_t\in\mathbb{R}^{N_z\times d_z}.$
We refer to $\boldsymbol{z}_t$ as a \emph{latent action}, which characterizes the observed transition between consecutive states. Here, $q$ denotes the posterior over the latent action, parameterized by the inverse dynamics model $\mathcal{I}$. We then use this latent action to model how the current state evolves into the subsequent state via the forward dynamics model $\mathcal{F}$, which predicts ${\boldsymbol{s}}_{t+1} \sim p_{\scriptscriptstyle\mathcal{F}} \left({\boldsymbol{s}}_{t+1} \,|\, \boldsymbol{s}_t, \boldsymbol{z}_t \right)$.

The inverse and forward dynamics therefore form a coupled transition-learning process: $\mathcal{I}$ identifies the transition from consecutive observations, while $\mathcal{F}$ models how that transition evolves the current state. Importantly, $\boldsymbol{z}_t$ is not required to recover the task-specific physical action executed between the two observations. Across tasks, these latent actions lie in a shared transition space $\mathcal{Z}$, allowing $\mathcal{F}$ to be shared without aligning the heterogeneous physical action spaces of VLN and EVT. Meanwhile, $\boldsymbol{s}_t^\tau$ remains task-specific, so sharing $\mathcal{F}$ does not require a unified state representation or assume identical state evolution across tasks.
To learn this coupled dynamics, we jointly optimize the inverse and forward models via a state prediction objective: 
\begin{equation}
\mathcal{L}_{\mathrm{state}}
=
-
\mathbb{E}_{
\boldsymbol{z}_t
\sim q
\left(
\boldsymbol{z}_t
\,|\,
\boldsymbol{h}_t,
\boldsymbol{h}_{t+1}
\right)
}
\left[
\log
p_{\scriptscriptstyle\mathcal{F}}
\left(
\boldsymbol{s}_{t+1}
\,|\,
\boldsymbol{s}_t,
\boldsymbol{z}_t
\right)
\right].
\label{eq:state_prediction_loss}
\end{equation}
This objective encourages the inferred latent action to {capture} the observed state change, so that $\boldsymbol{z}_t$ {encodes} transition information directly useful for predicting $\boldsymbol{s}_{t+1}$.

\noindent\textbf{Causal Latent-Action Prediction.}
The inverse dynamics model $\mathcal{I}$ infers latent action from consecutive observations, providing the latent transition used to learn the forward dynamics. However, this inference requires $\boldsymbol{h}_{t+1}=\mathrm{VLM}(\boldsymbol{o}_{\leq t+1},\boldsymbol{l})$, which is unavailable during inference. To enable prediction without future observations, {we learn a causal prior over the latent action, parameterized by $\mathcal C$, using only the observed history and current world state:}
$\hat{\boldsymbol{z}}_t \sim p\left(\hat{\boldsymbol{z}}_t \,|\, \boldsymbol{h}_{\leq t}, \boldsymbol{s}_t\right).$
Here, $\hat{\boldsymbol{z}}_t\in\mathbb{R}^{N_z\times d_z}$ denotes a causal latent-action sample. We then align the causal prior with the posterior via KL divergence~\citep{kullback1951information}:
\begin{equation}
\mathcal{L}_{\mathrm{KL}}
=
D_{\mathrm{KL}}
\left(
q
\left(
\boldsymbol{z}_t
\,|\,
\boldsymbol{h}_t,
\boldsymbol{h}_{t+1}
\right)
\;\Vert\;
p
\left(
{\boldsymbol{z}}_t
\,|\,
\boldsymbol{h}_{\leq t},
\boldsymbol{s}_t
\right)
\right),
\label{eq:causal_transition_loss}
\end{equation}
which transfers the transition information from the posterior to the causal prior, allowing the latent transition to be predicted without access to future observations. {Note that both distributions are defined over the random variable $\boldsymbol{z}_t$, while the hat notation $\hat{\boldsymbol{z}}_t$ is used specifically to distinguish the causal latent-action sample from the future-informed one.}

The KL objective constrains the causal prediction in latent-action space, but does not directly supervise the resulting state prediction. We therefore feed $\hat{\boldsymbol{z}}_t$ into the shared forward dynamics model $\mathcal{F}$, yielding $\hat{\boldsymbol{s}}_{t+1} \sim p_{\scriptscriptstyle\mathcal{F}}\left(\hat{\boldsymbol{s}}_{t+1} \,|\, \boldsymbol{s}_t, \hat{\boldsymbol{z}}_t\right)$, a causal estimate of the future state usable at decision time. Since $\boldsymbol{s}_{t+1}$ is a low-dimensional projection of $\boldsymbol{h}_{t+1}$, state-space supervision alone may be insufficient. To address this, we introduce a decoder $d(\cdot)$ that reconstructs $\boldsymbol{h}_{t+1}$ from $\hat{\boldsymbol{s}}_{t+1}$:
\begin{equation}
\mathcal{L}_{\mathrm{rec}}
=
\mathbb{E}_{
\hat{\boldsymbol{z}}_t
\sim
p\left(
\boldsymbol{z}_t
\mid
\boldsymbol{h}_{\leq t},
\boldsymbol{s}_t
\right)
}
\left[
\left\|
d\left(\hat{\boldsymbol{s}}_{t+1}\right)
-
\boldsymbol{h}_{t+1}
\right\|_2^2
\right].
\label{eq:representation_reconstruction}
\end{equation}
Together, these objectives supervise the dynamics learning at complementary
levels: $\mathcal{L}_{\mathrm{state}}$ learns the forward dynamics,
$\mathcal{L}_{\mathrm{KL}}$ aligns the causal prior with the posterior,
and $\mathcal{L}_{\mathrm{rec}}$ further constrains the resulting
future-state prediction through the future observation representation.
Then the overall dynamics-learning objective is:
\begin{equation}
\mathcal{L}_{\mathrm{Dyn}}
=
\mathcal{L}_{\mathrm{state}}
+
\mathcal{L}_{\mathrm{rec}}
+
\lambda\mathcal{L}_{\mathrm{KL}},
\label{eq:dynamics_objective}
\end{equation}
where $\lambda$ balances the KL term.
In practice, $\mathcal{I}$, $\mathcal{C}$, and $\mathcal{F}$ are implemented as Transformer-based recurrent modules.
Both $\mathcal{I}$ and $\mathcal{C}$ model probabilistic latent transitions as the future-informed posterior and causal prior, respectively, and are aligned via $\mathcal{L}_{\mathrm{KL}}$.
At inference, $\mathcal{I}$ and the decoder $d(\cdot)$ are discarded; only $\mathcal{C}$ and $\mathcal{F}$ are retained to predict $\hat{\bm{s}}_{t+1}$ without future observations.
Implementation details appear in \underline{Appendix\,\textbf{B.1}
of the Supplementary Material}.


\subsection{Adapting Shared Dynamics Across Tasks}
\label{sec:task_integration}
While the shared world dynamics predict how the current state may evolve, VLN and EVT retain distinct task representations and decision processes. We thus use a common context construction to incorporate the shared prediction, with task-specific readouts for adaptation.
Specifically, we construct a decision context $\boldsymbol{c}_t^\tau$ by concatenating the current observation representation, world state, and predicted future state. {For each task, a learnable query representation $\boldsymbol{q}_t^\tau$ is extracted from the VLM features and used to retrieve relevant information from this predictive context for downstream decision making.}
The task representation $\boldsymbol{q}_t^\tau$ then selectively retrieves relevant information from this context via a task-specific readout ($
\tau\in\{\mathrm{VLN},\mathrm{EVT}\}$):
\begin{equation}
\boldsymbol{c}_t^\tau
=
\operatorname{Concat}
\left(
\boldsymbol{h}_t^\tau,\,
\boldsymbol{s}_t^\tau,\,
\hat{\boldsymbol{s}}_{t+1}^\tau
\right), 
\boldsymbol{q}_t^{\tau,*}
=
\boldsymbol{q}_t^\tau
+
\mathcal{R}_\tau
\left(
\boldsymbol{q}_t^\tau,\,
\boldsymbol{c}_t^\tau
\right),
\label{eq:task_adaptation}
\end{equation}
Here, \(\mathcal{R}_\tau\) denotes a task-specific adaptation function, which we instantiate with cross-attention between \(\boldsymbol{q}_t^\tau\) and \(\boldsymbol{c}_t^\tau\). The shared context exposes both tasks to the predicted state evolution, while the residual connection preserves task-specific information. From $\boldsymbol{q}_t^{\tau,*}$, each task branch further incorporates the signals required for its own decision making.

\noindent\textbf{For VLN}, $\boldsymbol{q}_t^{\mathrm{VLN},*}$ incorporates the state evolution predicted by the shared world dynamics into the instruction-conditioned task representation. As navigation unfolds over an extended instruction-guided trajectory, the current decision can further benefit from recent navigation experience. We therefore augment the predictive representation with an action-aware episodic memory $\mathcal{M}_t=\{\boldsymbol{m}_i\}_{i<t}$, where each entry $\boldsymbol{m}_i$ encodes the previous task representation $\boldsymbol{q}_i^{\mathrm{VLN}}$, the executed action $\boldsymbol{a}_i^{\mathrm{VLN}}$, and its temporal distance to the current step. By associating previous task representations with the actions executed from them, the memory provides historical state--action context for the current decision rather than merely caching past observations.

While the episodic memory complements the predictive representation with past experience, action supervision alone focuses on the immediate decision and offers limited constraint on how the agent's motion evolves over subsequent steps. We therefore introduce an auxiliary future-motion objective that predicts the agent's $K$-step ego-trajectory, providing explicit multi-step supervision that encourages the predictive representation to retain information beyond that required for the current action. Accordingly, the VLN predictions are formulated as
\begin{equation}
\begin{aligned}
p \big(
\boldsymbol{a}_t^{\mathrm{VLN}},
&\boldsymbol{\xi}_{t}^{\mathrm{VLN}}
\,|\,
\boldsymbol{q}_t^{\mathrm{VLN},*},
\mathcal{M}_t
\big)
= \\
&p \big(
\boldsymbol{a}_t^{\mathrm{VLN}}
\,|\,
\boldsymbol{q}_t^{\mathrm{VLN},*},
\mathcal{M}_t
\big)\,
p \big(
\boldsymbol{\xi}_{t}^{\mathrm{VLN}}
\,|\,
\boldsymbol{q}_t^{\mathrm{VLN},*}
\big),
\end{aligned}
\label{eq:vln_adaptation}
\end{equation}
where $\boldsymbol{a}_t^{\mathrm{VLN}}=(u_t,r_t)$, with
$u_t\!\in\!\{\textsc{Stop},\textsc{Forward},$ $\textsc{Left},\textsc{Right}\}$
denoting the action type and $r_t$ its magnitude, namely movement
distance for \textsc{Forward} and turning angle for \textsc{Left} / \textsc{Right}. $\boldsymbol{\xi}_t^{\mathrm{VLN}}\!\in\!\mathbb{R}^{K\times3}$ denotes the $K$-step future ego-trajectory predicted at time $t$, with each ego-pose represented by two-dimensional planar position and heading in the current agent-centric coordinate frame. The two factors serve distinct roles: episodic memory directly informs current action prediction, whereas trajectory prediction provides task-specific multi-step motion supervision. Notably, the trajectory does not condition the navigation action and serves only as an auxiliary training objective.

\noindent\textbf{For EVT}, translating the predicted future state into a tracking action requires reasoning about its implications for agent--target interaction. Unlike VLN, which benefits from episodic memory to retain navigation experience along an extended instruction-guided trajectory, EVT primarily requires responsive control with respect to a continuously moving target. Recent motion cues are already captured by its consecutive visual observations; we therefore do not introduce additional episodic memory, but instead focus the task adaptation on the current agent--target relation and anticipated ego-motion.
Specifically, whereas VLN uses future ego-trajectory prediction as auxiliary supervision, EVT models a progressive dependency among target localization, future ego-motion, and action prediction. We formulate their joint predictive distribution as:
\begin{equation}
\begin{aligned}
&p\left(
\boldsymbol{a}_t^{\mathrm{EVT}},
\boldsymbol{\xi}_t^{\mathrm{EVT}},
\boldsymbol{x}_t^{\mathrm{EVT}}
\mid \boldsymbol{q}_t^{\mathrm{EVT},*}
\right)\\
&= p\left(
\boldsymbol{x}_t^{\mathrm{EVT}}
\mid \boldsymbol{q}_t^{\mathrm{EVT},*}
\right)
p\left(
\boldsymbol{\xi}_t^{\mathrm{EVT}}
\mid \boldsymbol{x}_t^{\mathrm{EVT}},
\boldsymbol{q}_t^{\mathrm{EVT},*}
\right)\\
&\quad \cdot
p\left(
\boldsymbol{a}_t^{\mathrm{EVT}}
\mid \boldsymbol{\xi}_t^{\mathrm{EVT}},
\boldsymbol{q}_t^{\mathrm{EVT},*}
\right),
\end{aligned}
\label{eq:evt_factorization}
\end{equation}
where $\boldsymbol{x}_t^{\mathrm{EVT}}\in\mathbb{R}^{2}$ denotes the target position in the agent-centric frame, $\boldsymbol{\xi}_t^{\mathrm{EVT}}\in\mathbb{R}^{K\times2}$ the predicted $K$-step ego-trajectory (each entry a planar position), and $\boldsymbol{a}_t^{\mathrm{EVT}}=(v_t,\omega_t)\in\mathbb{R}^{2}$ the linear and angular velocities. We implement these dependencies through gated residual integration, where each intermediate prediction is selectively incorporated into the representation for the subsequent stage. Detailed formulations are provided in \underline{Appendix\,\textbf{B.2}}.

\subsection{Two-Stage Joint Learning}
\label{sec:training}

We train \textup{\textsc{WorldWeave}} in two stages: shared dynamics learning and task adaptation. In the first stage, we learn shared predictive dynamics from heterogeneous real-world and simulated videos without requiring physical-action annotations, enabling the model to capture state-transition patterns across diverse embodied interactions. In the second stage, the learned dynamics are transferred to VLN and EVT through lightweight task-specific adaptation using embodied task data. This two-stage design separates general dynamics learning from downstream task adaptation, allowing the shared dynamics to be reused while preserving the distinct state representations and decision processes of each task. Details of data composition and training configurations are given in \underline{Appendix\,\textbf{A} of Supplemenary Material}.

\noindent\textbf{Stage I: Shared Dynamics Learning.}
We first learn the shared world dynamics introduced in Sec.~\ref{sec:shared_dynamics} from heterogeneous real-world and simulated videos, without requiring physical-action annotations.This stage jointly adapts the pretrained vision-language backbone~\citep{bai2025qwen3} and learns the dynamics modules using $\mathcal{L}_{\mathrm{Dyn}}$ in Eq.(\ref{eq:dynamics_objective}), which optimizes state prediction, representation reconstruction, and latent-action alignment.
During training, latent actions are sampled via reparameterization from the posterior and causal prior distributions, while the KL divergence aligns the causal prior with the posterior. The posterior latent action is used to learn the shared forward dynamics $\mathcal{F}$, while the causal latent action is used to reconstruct the future observation representation.
After Stage I, the inverse dynamics model $\mathcal{I}$ and representation decoder $d$ are discarded. The causal latent-action predictor $\mathcal{C}$ and shared forward dynamics model $\mathcal{F}$ are retained to predict future states using only information available at the current step.

\noindent\textbf{Stage II: Task Adaptation.}
We incorporate the pretrained world dynamics into VLN and EVT through the task-specific adaptation described in Sec.~\ref{sec:task_integration}. The dynamics modules are frozen, while the vision-language backbone is adapted via separate LoRA adapters for each task, along with their respective action heads. Although both tasks share the same predictive dynamics, their learning objectives reflect their distinct decision processes.

For VLN, we jointly supervise the navigation action
$\boldsymbol{a}_t^{\mathrm{VLN}}$ and the future
ego-trajectory $\boldsymbol{\xi}_t^{\mathrm{VLN}}$:
\begin{equation}
\begin{aligned}
\mathcal{L}_{\mathrm{VLN}}
= \mathbb{E}\Big[
-&\log p\big(
\boldsymbol{a}_t^{\mathrm{VLN,gt}}
\mid
\boldsymbol{q}_t^{\mathrm{VLN},*},
\mathcal{M}_t
\big)\\
&+\ell\big(
\boldsymbol{\xi}_t^{\mathrm{VLN}},
\boldsymbol{\xi}_t^{\mathrm{VLN,gt}}
\big)
\Big],
\end{aligned}
\label{eq:vln_loss}
\end{equation}
where the superscript $\mathrm{gt}$ denotes ground-truth supervision. We use $\ell_1$ loss for trajectory regression $\ell(\cdot,\cdot)$ and categorical cross-entropy for both action type and motion parameters. The trajectory loss is computed over valid future steps and serves only as auxiliary supervision.

For EVT, the task adaptation successively predicts the agent-centric target
position $\boldsymbol{x}_t^{\mathrm{EVT}}$, future ego-trajectory
$\boldsymbol{\xi}_t^{\mathrm{EVT}}$, and action
$\boldsymbol{a}_t^{\mathrm{EVT}}=(v_t,\omega_t)$. We jointly supervise
these predictions as
\begin{equation}
\begin{aligned}
\mathcal{L}_{\mathrm{EVT}}
= \mathbb{E}\Big[
&\ell\big(
\boldsymbol{x}_t^{\mathrm{EVT}},
\boldsymbol{x}_t^{\mathrm{EVT,gt}}
\big)+\ell\big(
\boldsymbol{\xi}_t^{\mathrm{EVT}},
\boldsymbol{\xi}_t^{\mathrm{EVT,gt}}
\big)\\
&+\ell\big(
\boldsymbol{a}_t^{\mathrm{EVT}},
\boldsymbol{a}_t^{\mathrm{EVT,gt}}
\big)
\Big].
\end{aligned}
\label{eq:evt_loss}
\end{equation}

\section{Experiments}\label{sec:exper}
We evaluate \textup{\textsc{WorldWeave}} on VLN and EVT tasks through comparisons with existing methods, ablations of cross-task dynamics sharing and task-specific adaptation, and real-world robot deployment. 
\textbf{More details, experiments, and further analyses are given in \underline{Appendix}.}

\noindent\textbf{Datasets.}
We evaluate on \texttt{R2R-CE}\,\citep{r2r,Beyondthenavgraph}, \texttt{RxR-CE}\,\citep{rxr}, 
\texttt{VLN-PE}\,\citep{vlnpe}, 
and \texttt{VLNVerse}\,\citep{vlnverse} for VLN, 
and \texttt{EVT-}
\texttt{Bench}\,\citep{trackvla}, 
\texttt{UnrealZoo}\,\citep{unrealzoo}, and \texttt{Gym-UnrealCV}\,\citep{unrealcv} for EVT. Training details are provided in \underline{Appendix\,\textbf{A}}.


\noindent\textbf{Evaluation Metrics.}
For VLN, we report Success Rate (SR), Success weighted by Path Length (SPL), Navigation Error (NE), Normalized Dynamic Time Warping (NDTW), and Success weighted by NDTW (SDTW)~\citep{Visionandlanguagenavigation}. For EVT, we report Success Rate (SR), Tracking Rate (TR), and Collision Rate (CR). Details are provided in \underline{Appendix\,\textbf{B.3}}.

\begin{table}[t]
\centering
\caption{Cross-task performance comparison of task-specific baselines, Uni-NaVid, and \textup{\textsc{WorldWeave}} on VLN and EVT tasks.}
\vspace{-0.2cm}
\label{tab:cross_task}
\resizebox{\linewidth}{!}{
\begin{tabular}{lcccccc}
\toprule
& \multicolumn{2}{c}{\textbf{R2R-CE}}
& \multicolumn{2}{c}{\textbf{RxR-CE}}
& \multicolumn{2}{c}{\textbf{EVT-Bench-STT}} \\
\cmidrule(lr){2-3}
\cmidrule(lr){4-5}
\cmidrule(lr){6-7}
\textbf{Method}
& \textbf{SR}$\uparrow$
& \textbf{SPL}$\uparrow$
& \textbf{SR}$\uparrow$
& \textbf{SPL}$\uparrow$
& \textbf{SR}$\uparrow$
& \textbf{TR}$\uparrow$ \\
\midrule

Baseline (VLN)
& 46.9 & 41.9
& 45.8 & 38.1
& -- & -- \\

Baseline (EVT)
& -- & --
& -- & --
& 41.1 & 71.2 \\

Uni-NaVid~\citep{zhang2024uni}
& 47.0 & 42.7
& 48.7 & 40.9
& 25.7 & 39.5 \\

\rowcolor{mypurple!20}
\textbf{\textsc{WorldWeave}}
& \textbf{54.2} & \textbf{49.4}
& \textbf{52.7} & \textbf{45.9}
& \textbf{78.1} & \textbf{88.0} \\

\bottomrule
\end{tabular}}\vspace{-0cm}
\end{table}

\noindent\textbf{Cross-Task Overview.}
Before examining each task individually, we provide a cross-task comparison to assess the overall performance of \textup{\textsc{WorldWeave}}. 
We compare it with independently trained task-specific baselines and Uni-NaVid~\citep{zhang2024uni}, a publicly available generalist model that supports both instruction-guided navigation and human following. Other recent generalist models covering navigation and target tracking have not publicly released their code or model weights, and are therefore discussed in the corresponding task-specific comparisons.
As shown in Table~\ref{tab:cross_task}, \textup{\textsc{WorldWeave}} improves over the corresponding task-specific baselines on both VLN and EVT, while also outperforming Uni-NaVid across all evaluated settings.


\begin{table*}[t]
\centering
\vspace{-0.1cm}
\caption{
Comparison on R2R-CE and RxR-CE Val-Unseen splits under monocular settings. Results outperforming the baseline are \textbf{bold}; best results among publicly available methods are \underline{underlined}. \textit{\textcolor{Gray}{{Gray rows}} indicate recent VLM-based methods
whose code or training data are not publicly released; we include their
officially reported results for reference only.}
}
\label{tab:vln-main}
\vspace{-0.1cm}
\resizebox{0.96\textwidth}{!}{
\begin{threeparttable}
\begin{tabular}{@{}l|ccccc|cccccc@{}}
\toprule
& \multicolumn{5}{c|}{\textbf{R2R-CE Val-Unseen}}
& \multicolumn{6}{c}{\textbf{RxR-CE Val-Unseen}} \\
\textbf{Methods}
& TL $\downarrow$
& NE $\downarrow$
& OSR $\uparrow$
& SR $\uparrow$
& SPL $\uparrow$
& TL $\downarrow$
& NE $\downarrow$
& OSR $\uparrow$
& SR $\uparrow$
& SPL $\uparrow$
& NDTW $\uparrow$ \\
\midrule


monoVLN~\citep{Lu2025monoVLN}
& - & \underline{4.61} & 62.4 & 54.8 & 44.4
& - & 8.29 & 37.7 & 31.8 & 26.8 & - \\

SC$^2$-WM~\citep{yao2026sc}
& 17.65 & 5.37 & 58.8 & 50.9 & 37.2
& 20.87 & 8.36 & 48.4 & 35.8 & 27.2 & 44.9 \\


{UniNaVid (Vicuna-7B)~\citep{zhang2024uni}}
& {-}
& {5.58}
& {53.3}
& {47.0}
& {42.7}
& {-}
& {6.24}
& {-}
& {48.7}
& {40.9}
& {-} \\

NaVILA (VILA-1.5-3B)~\citep{cheng2024navila}
& - & 5.43 & 62.5 & 54.0 & 49.0
& - & 6.77 & - & 49.3 & 44.0 & 58.8 \\

StreamVLN (Qwen2-7B)~\citep{wei2025streamvln}
& - & 4.98 & \underline{64.2} & 56.9 & 51.9
& - & 6.22 & - & 52.9 & 46.0 & 61.9 \\

InternVLA-N1 (Qwen2.5-7B)~\citep{internvlan1}
& - & {4.83} & 63.3 & \underline{58.2} & \underline{54.0}
& - & \underline{5.91} & - & \underline{53.5} & \underline{46.1} & \underline{65.3} \\

\textcolor{Gray}{QwenRobotNav (Qwen3-VL-4B)~\citep{qwenrobotnav}}
& \textcolor{Gray}{-}
& \textcolor{Gray}{4.22}
& \textcolor{Gray}{73.6}
& \textcolor{Gray}{66.9}
& \textcolor{Gray}{60.5}
& \textcolor{Gray}{-}
& \textcolor{Gray}{4.15}
& \textcolor{Gray}{-}
& \textcolor{Gray}{71.3}
& \textcolor{Gray}{61.5}
& \textcolor{Gray}{68.6} \\

\cmidrule(lr){1-12}

Baseline (Qwen3-VL-4B)~\citep{bai2025qwen3vl}
& \underline{9.00} & 6.20 & 52.1 & 46.9 & 41.9
& \underline{11.29} & 7.24 & 45.8 & 41.2 & 38.1 & 55.2 \\

\rowcolor{mypurple!20}
\textbf{\textup{\textsc{WorldWeave}} (Qwen3-VL-4B)}
& 11.3
& \textbf{5.78}
& \textbf{61.5}
& \textbf{54.2}
& \textbf{49.4}
& 13.89
& \textbf{6.31}
& \underline{\textbf{52.7}} & {\textbf{50.1}}
& \textbf{45.9}
& \textbf{62.9} \\

\bottomrule
\end{tabular}
\end{threeparttable}
}
\vspace{0cm}
\end{table*}

\begin{table*}[t]
\vspace{-0.1cm}
    \caption{
    {Comparison on EVT-Bench.} 
     Results outperforming the baseline are \textbf{bold}. }
     \vspace{-0.2cm}
    \centering
    \resizebox{0.96\textwidth}{!}{
    \begin{threeparttable}
    \begin{tabular}{lccccccccc}
        \toprule
        \multirow{2}{*}{Methods}& \multicolumn{3}{c}{\textit{Single-Target Tracking (STT)}} & \multicolumn{3}{c}{\textit{Distracted Tracking (DT)}} & \multicolumn{3}{c}{\textit{Ambiguity Tracking (AT)}} \\
        \cmidrule(lr){2-4} \cmidrule(lr){5-7} \cmidrule(lr){8-10}
        & SR$\uparrow$ & TR$\uparrow$ & CR$\downarrow$ & SR$\uparrow$ & TR$\uparrow$ & CR$\downarrow$ & SR$\uparrow$ & TR$\uparrow$ & CR$\downarrow$ \\
        \midrule
EVT~\citep{zhong2024EVT} & 32.5 & 49.9 & 40.5 & 15.7 & 35.7 & 53.3 & 18.3 & 21.0 & 44.9 \\
Uni-NaVid (Vicuna-7B)~\citep{zhang2024uni} & 25.7 & 39.5 & 41.9 & 11.3 & 27.4 & 43.5 & 8.26 & 28.6 & 43.7 \\
{OpenTrackVLA (Qwen-0.6B)~\citep{opentrackvla2025}}  & 47.2 & 58.7 & 5.72 & 22.5 & 42.8 & 8.93 & 28.6 & 51.8 & {6.37} \\
\textcolor{mygray}{Qwen-RobotNav (Qwen3-VL-4B)~\citep{qwenrobotnav}} & \textcolor{mygray}{77.4} & \textcolor{mygray}{90.0} & \textcolor{mygray}{6.40} & \textcolor{mygray}{-} & \textcolor{mygray}{-} & \textcolor{mygray}{-} & \textcolor{mygray}{-} & \textcolor{mygray}{-} & \textcolor{mygray}{-} \\
\textcolor{mygray}{NavFoM(Qwen2-7B)~\citep{zhang2026embodied}} & \textcolor{mygray}{85.0} & \textcolor{mygray}{80.5} & \textcolor{mygray}{-} & \textcolor{mygray}{61.4} & \textcolor{mygray}{68.2} & \textcolor{mygray}{-} & \textcolor{mygray}{-} & \textcolor{mygray}{-} & \textcolor{mygray}{-} \\
\textcolor{mygray}{TrackVLA++ (Qwen2-7B)(\cite{trackvla++})} & \textcolor{mygray}{86.0} & \textcolor{mygray}{81.0} & \textcolor{mygray}{2.10} & \textcolor{mygray}{66.5} & \textcolor{mygray}{68.8} & \textcolor{mygray}{4.71} & \textcolor{mygray}{51.2} & \textcolor{mygray}{63.4} & \textcolor{mygray}{15.9} \\

        \midrule

        {Baseline (Qwen3-VL-4B)\citep{bai2025qwen3vl}} & 41.1 & 71.2 & 6.50 & 16.3 & 45.7 & 19.4 & 14.8 & 47.1 & 8.70  \\
        \rowcolor{mypurple!20}
        \textbf{\textup{\textsc{WorldWeave}} (Qwen3-VL-4B)} & {\textbf{78.1}} & {\textbf{88.0}} & \textbf{4.89} & {\textbf{41.7}} & {\textbf{75.2}} & \textbf{14.24} & {\textbf{51.2}} & \textbf{74.5} &  18.62
        \\
        \bottomrule
    \end{tabular}
    \begin{tablenotes}    
	\footnotesize                              
	\item Note: Evaluation metrics are defined as follows: \textbf{Success Rate (SR)}, proportion of episodes that the agent ends correctly oriented within 1–3m of the target; \textbf{Tracking Rate (TR)}, proportion of timesteps with successful target tracking; and \textbf{Collision Rate (CR)}, proportion of episodes terminated due to collisions. 
	\end{tablenotes}           
	\end{threeparttable}
    \vspace{0cm}
    }
    \label{tab:evt-bench}
\end{table*}

\subsection{Comparison with State-Of-The-Art Methods}
\label{sec:vln_results}
\noindent\textbf{Vision-and-Language Navigation.}
Table~\ref{tab:vln-main} compares our method with existing methods on \texttt{R2R-CE} and \texttt{RxR-CE} under monocular settings.
On \texttt{R2R-CE}, \textup{\textsc{WorldWeave}} improves the matched Qwen3-VL baseline from 46.9 to 54.2 in SR and from 41.9 to 49.4 in SPL, while reducing NE from 6.20 to 5.78.
The gains transfer to the more instruction-diverse \texttt{RxR-CE}, where SR\,/\,SPL increase from 45.8\,/\,38.1 to 52.7\,/\,45.9, with improved NDTW further indicating better trajectory-level navigation quality.
These improvements are consistent with our VLN-specific adaptation, where action-aware episodic memory retains navigation experience and future ego-trajectory prediction provides multi-step motion supervision.
While some recent VLM-based methods achieve higher absolute performance, \textup{\textsc{WorldWeave}} consistently improves the matched task-specific baseline and remains competitive with publicly available VLN methods.
Consistent gains on \texttt{VLN-PE} and \texttt{VLNVerse} are further reported in the \underline{Appendix\,\textbf{C.1}}.


\begin{table*}[t]
\centering
\vspace{0cm}
\caption{
Comparison of cross-task sharing strategies on R2R Val-Unseen and EVT-Bench STT sets.
Variants (a)--(d) progressively examine the sharing of state representations,
predictive dynamics, and task-specific predictive adaptation.
Best results are in \textbf{bold}.
}
\label{tab:sharing_strategy}
\vspace{-0.2cm}
\resizebox{\textwidth}{!}{
\begin{tabular}{lccc ccc ccc}
\toprule
& \multicolumn{3}{c}{\textbf{Model Design}}
& \multicolumn{3}{c}{\textbf{R2R Val-Unseen}}
& \multicolumn{3}{c}{\textbf{EVT-Bench STT}} \\
\cmidrule(lr){2-4}
\cmidrule(lr){5-7}
\cmidrule(lr){8-10}
\textbf{Model Design}
& \textbf{State Representation}
& \textbf{Dynamics}
& \textbf{Predictive Adaptation}
& \textbf{OSR} $\uparrow$
& \textbf{SR} $\uparrow$
& \textbf{SPL} $\uparrow$
& \textbf{SR} $\uparrow$
& \textbf{TR} $\uparrow$
& \textbf{CR} $\downarrow$ \\
\midrule
\multicolumn{10}{l}{\textit{\textcolor{gray}{Single-Task Training}}} \\
Baseline (VLN Only)
& Task-specific & -- & --
& 52.1 &46.9 &41.9
 & -- & -- & -- 
\\
\rowcolor{mygray!15}
Ind-WM (VLN Only)
& Task-specific & Independent $\mathcal{C},\mathcal{F}$ & --
& 56.2 &48.1 &45.7
 & -- & -- & -- 
\\

Baseline (EVT Only)
& Task-specific & -- & --
& -- & -- & --
& 41.1 & 71.2 & 6.50\\
\rowcolor{mygray!15}
Ind-WM (EVT Only)
& Task-specific & Independent $\mathcal{C},\mathcal{F}$ & --
& -- & -- & --
& 60.1 & 72.3 & 8.41\\

\midrule
\multicolumn{10}{l}{\textit{\textcolor{gray}{Cross-Task Joint Training}}} \\

(a)
& Shared & -- & --
& 57.7
& 50.2
& 46.3 & 29.1 & 58.6
& 25.1  \\

(b)
& Task-specific& -- & --
& 46.3
& 41.8
& 39.1 & 68.0 & 77.1
& \textbf{3.41}  \\

(c)
& Task-specific
& Shared $\mathcal{C},\mathcal{F}$
& Shared
& 55.2 & 48.5 & 44.3
& 71.2 & 72.7 & 4.82  \\

\rowcolor{mypurple!20}
\textbf{(d) \textup{\textsc{WorldWeave}}}
& Task-specific
& Shared $\mathcal{C},\mathcal{F}$
& Task-specific $\mathcal{R}_{\tau}$
& \textbf{61.5} & \textbf{54.2} & \textbf{49.4} & \textbf{78.1} & \textbf{88.0} & {4.89} \\

\bottomrule
\end{tabular}}
\vspace{0cm}
\end{table*}

\noindent\textbf{Embodied Visual Tracking.}
On \texttt{EVT-Bench}, \textup{\textsc{WorldWeave}} substantially improves over matched Qwen3-VL baseline in SR and TR (Table~\ref{tab:evt-bench}), yielding particularly notable gains in more challenging DT and AT settings involving distractors and target ambiguity. These improvements align with our EVT-specific adaptation, which progressively integrates predicted target position and ego-motion to tailor shared predictions to agent--target interactions. While Collision Rate (CR) decreases under STT and DT, it increases under AT (8.70 to 18.62), despite a substantial improvement in SR (14.8 to 51.2). 
This may be partly because \textup{\textsc{WorldWeave}} maintains target tracking for more timesteps, increasing the agent's exposure to potential collisions in ambiguous scenes.
While some recent methods shown in gray achieve higher absolute performance, \textup{\textsc{WorldWeave}} consistently improves the matched baseline and remains competitive among publicly available methods. Additional evaluations on \texttt{UnrealZoo} and \texttt{Gym-UnrealCV} further show that tracking gains extend to changing targets and zero-shot conditions (please refer to \underline{Appendix\,\textbf{C.2}}).



\subsection{Ablation Studies}
\label{sec:ablation}
\vspace{-0.1CM}
\noindent\textbf{What Should Be Shared Across Tasks?}
A central design choice in \textup{\textsc{WorldWeave}} is to share predictive dynamics while preserving task-specific representations and adaptation. We first establish two single-task references: the task-specific baselines without explicit world modeling (Baseline), and their corresponding variants augmented with independently trained task-specific dynamics models (Ind-WM), which incorporate $\mathcal{C}$ and $\mathcal{F}$ without cross-task parameter sharing. We then examine four joint-training variants with different sharing boundaries:
\textbf{(a)} a shared LoRA adapter,
\textbf{(b)} task-specific LoRA adapters,
\textbf{(c)} task-specific LoRA adapters combined with shared causal latent-action predictor $\mathcal{C}$ and forward dynamics model $\mathcal{F}$,
and \textbf{(d) \textup{\textsc{WorldWeave}}}, which further incorporates task-specific predictive adaptation $\mathcal{R}_{\tau}$.
As shown in Table~\ref{tab:sharing_strategy}, independently trained world models (Ind-WM) improve both tasks over their respective baselines, demonstrating the benefits of predictive modeling even without cross-task sharing. For VLN, Ind-WM improves SR\,/\,SPL from 46.9\,/\,41.9 to 48.1\,/\,45.7. For EVT, it substantially improves SR from 41.1 to 60.1, while also increasing TR from 71.2 to 72.3. 
These results establish the benefits of predictive modeling within individual tasks, motivating further investigation into cross-task knowledge sharing.

Among the joint-training variants, (a) favors VLN but severely degrades EVT (\textit{e.g.}, SR\,/\,CR to 29.1\,/\,25.1), while (b) restores EVT performance at the cost of VLN. These suggest that sharing task representations alone does not guarantee beneficial cross-task transfer. In contrast, adding shared $\mathcal{C}$ and $\mathcal{F}$ in (c), while retaining task-specific representations, improves the success rates of both tasks over (b), supporting the potential of shared predictive dynamics as a basis for cross-task knowledge transfer. Finally, (d) further improves upon (c) through task-specific predictive adaptation $\mathcal{R}_{\tau}$, achieving the best VLN performance and the highest EVT tracking success and tracking ratio. 
Compared with independently trained world models, the complete \textup{\textsc{WorldWeave}} design achieves higher EVT SR and TR while simultaneously reducing CR, demonstrating improved tracking performance without increased collision rates. 
These results highlight the effectiveness of combining shared predictive dynamics with task-specific predictive adaptation to leverage cross-task knowledge while accommodating distinct task representations and decision requirements.

\begin{figure}[t]
    \centering
    \includegraphics[width=\linewidth]{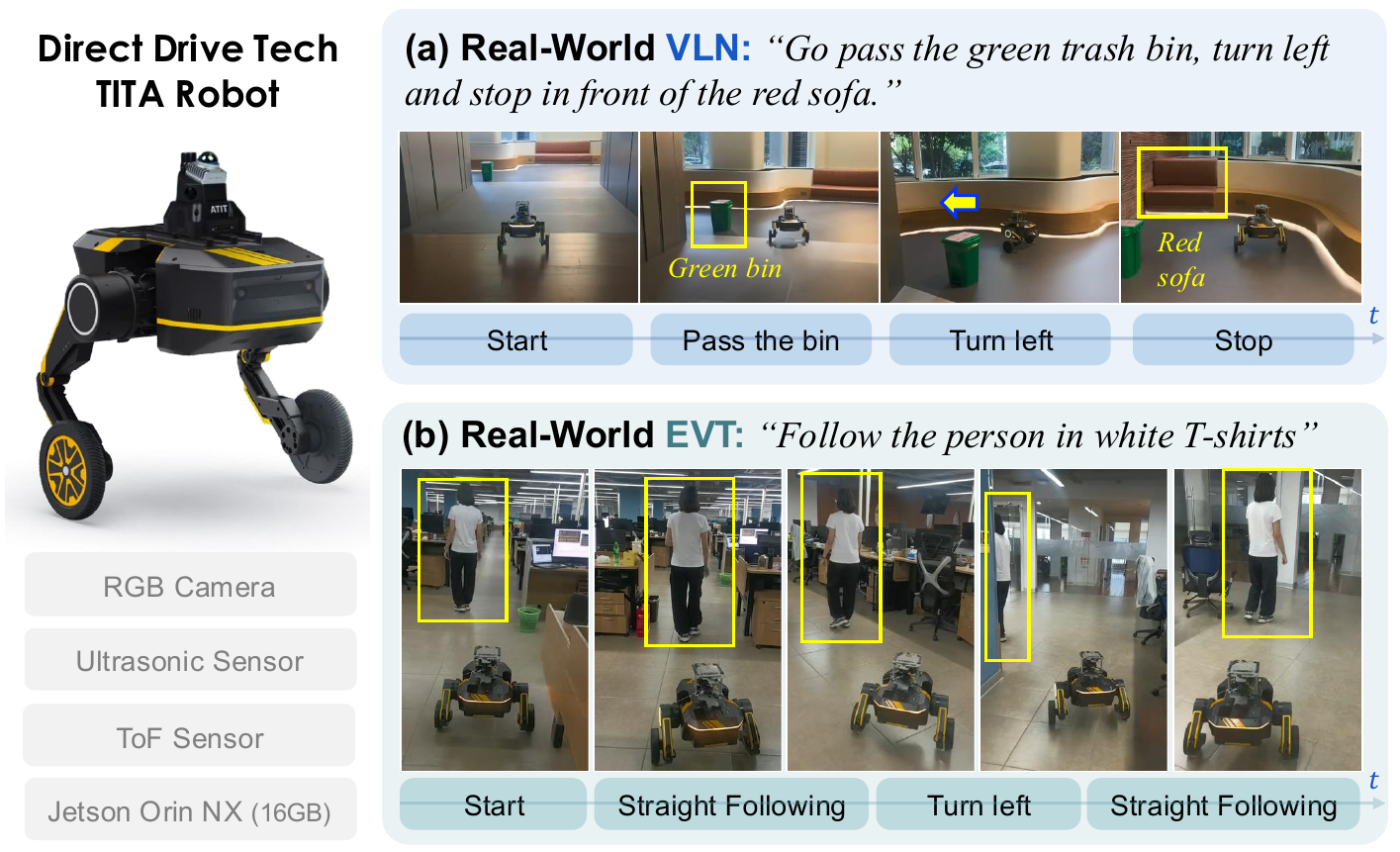}
    \vspace{-0.5cm}
    \caption{Real-world deployment for embodied navigation tasks.}
    \label{fig:realworld}
    \vspace{0cm}
\end{figure}

\subsection{Discussion}
\label{sec:discuss}\vspace{-0.1cm}
Recent methods in the \textcolor{mygray}{gray rows} achieve higher absolute performance on some metrics, but at substantially larger training scales ($\sim$8M--15.6M samples versus 500 hours of video and 2.5M task samples). We therefore focus on the controlled question of \emph{what should be shared across heterogeneous embodied tasks}. Our results suggest that sharing representations induces cross-task trade-offs, whereas sharing predictive dynamics with task-specific adaptation improves both VLN and EVT over their single-task baselines. This provides evidence that predictive dynamics can be shared without requiring unified task representations or decision processes.
More importantly, the design naturally lends itself to scaling: Stage-I can exploit heterogeneous videos without unified action annotations, while Stage-II accommodates task-specific adaptation. Due to computational constraints, we leave larger-scale training to future work, which will investigate whether such scaling further improves transferability, particularly to distinct embodied interactions such as manipulation.
\subsection{Real-World Robot Experiments}
\label{sec:real_robot}
We further deploy \textup{\textsc{WorldWeave}} on a TITA wheel-legged robot for both VLN and EVT in previously unseen indoor environments. As shown in Fig.~\ref{fig:realworld}, the robot performs closed-loop navigation and tracking under progressively challenging instructions and target-motion patterns, achieves 66.7\% and 66.0\% SR on VLN and EVT,
respectively, compared with 46.0\% and 48.7\% for the matched baselines. Detailed task protocols, and results are provided in \underline{Appendix\,\textbf{D} of Supplemenary Material}.

%



%


\section{Conclusion}\label{sec:conclu}
We introduced \textup{\textsc{WorldWeave}} to investigate shared predictive dynamics across heterogeneous embodied tasks, using VLN and EVT as complementary testbeds. By coupling shared dynamics with task-specific adaptation, our method consistently outperforms single-task baselines. These results demonstrate the potential of cross-task world modeling for learning reusable dynamics while accommodating distinct decision requirements.
Our findings offer a different perspective on embodied navigation: rather than learning task-specific dynamics independently, related embodied tasks can provide complementary interaction experiences for learning shared predictive dynamics. As embodied models expand to diverse tasks, key questions arise: which components can be shared, and how should shared predictions be adapted to different interaction patterns? Extending this approach beyond mobile embodied tasks to scenarios such as manipulation remains an important future direction. We will release our code and models upon acceptance.



{
    \small
    \bibliographystyle{ieeenat_fullname}
    \bibliography{main}

\begin{thebibliography}{55}
\providecommand{\natexlab}[1]{#1}
\providecommand{\url}[1]{\texttt{#1}}
\expandafter\ifx\csname urlstyle\endcsname\relax
  \providecommand{\doi}[1]{doi: #1}\else
  \providecommand{\doi}{doi: \begingroup \urlstyle{rm}\Url}\fi

\bibitem[Anderson et~al.(2018{\natexlab{a}})Anderson, Wu, Teney, Bruce,
  Johnson, S{\"u}nderhauf, Reid, Gould, and Van
  Den~Hengel]{Visionandlanguagenavigation}
Peter Anderson, Qi Wu, Damien Teney, Jake Bruce, Mark Johnson, Niko
  S{\"u}nderhauf, Ian Reid, Stephen Gould, and Anton Van Den~Hengel.
\newblock Vision-and-language navigation: Interpreting visually-grounded
  navigation instructions in real environments.
\newblock In \emph{CVPR}, pages 3674--3683. IEEE, 2018{\natexlab{a}}.

\bibitem[Anderson et~al.(2018{\natexlab{b}})Anderson, Wu, Teney, Bruce,
  Johnson, S{\"u}nderhauf, Reid, Gould, and Van Den~Hengel]{r2r}
Peter Anderson, Qi Wu, Damien Teney, Jake Bruce, Mark Johnson, Niko
  S{\"u}nderhauf, Ian Reid, Stephen Gould, and Anton Van Den~Hengel.
\newblock Vision-and-language navigation: Interpreting visually-grounded
  navigation instructions in real environments.
\newblock In \emph{CVPR}, pages 3674--3683. IEEE, 2018{\natexlab{b}}.

\bibitem[Bai et~al.(2025{\natexlab{a}})Bai, Cai, Chen, Chen, Chen, Cheng, Deng,
  Ding, Gao, Ge, et~al.]{bai2025qwen3}
Shuai Bai, Yuxuan Cai, Ruizhe Chen, Keqin Chen, Xionghui Chen, Zesen Cheng,
  Lianghao Deng, Wei Ding, Chang Gao, Chunjiang Ge, et~al.
\newblock Qwen3-vl technical report.
\newblock \emph{arXiv preprint arXiv:2511.21631}, 2025{\natexlab{a}}.

\bibitem[Bai et~al.(2025{\natexlab{b}})Bai, Cai, Chen, Chen, Chen, Cheng, Deng,
  Ding, Gao, Ge, et~al.]{bai2025qwen3vl}
Shuai Bai, Yuxuan Cai, Ruizhe Chen, Keqin Chen, Xionghui Chen, Zesen Cheng,
  Lianghao Deng, Wei Ding, Chang Gao, Chunjiang Ge, et~al.
\newblock Qwen3-vl technical report.
\newblock \emph{arXiv preprint arXiv:2511.21631}, 2025{\natexlab{b}}.

\bibitem[Bjorck et~al.(2025)Bjorck, Casta{\~n}eda, Cherniadev, Da, Ding, Fan,
  Fang, Fox, Hu, Huang, et~al.]{bjorck2025gr00t}
Johan Bjorck, Fernando Casta{\~n}eda, Nikita Cherniadev, Xingye Da, Runyu Ding,
  Linxi Fan, Yu Fang, Dieter Fox, Fengyuan Hu, Spencer Huang, et~al.
\newblock Gr00t n1: An open foundation model for generalist humanoid robots.
\newblock \emph{arXiv preprint arXiv:2503.14734}, 2025.

\bibitem[Bruce et~al.(2024)Bruce, Dennis, Edwards, Parker-Holder, Shi, Hughes,
  Lai, Mavalankar, Steigerwald, Apps, et~al.]{genie}
Jake Bruce, Michael~D Dennis, Ashley Edwards, Jack Parker-Holder, Yuge Shi,
  Edward Hughes, Matthew Lai, Aditi Mavalankar, Richie Steigerwald, Chris Apps,
  et~al.
\newblock Genie: Generative interactive environments.
\newblock In \emph{ICML}, 2024.

\bibitem[Chen et~al.(2024)Chen, Lin, Xu, Chai, Liang, and Wong]{mapgpt}
Jiaqi Chen, Bingqian Lin, Ran Xu, Zhenhua Chai, Xiaodan Liang, and Kwan-Yee
  Wong.
\newblock Mapgpt: Map-guided prompting with adaptive path planning for
  vision-and-language navigation.
\newblock In \emph{Proceedings of the 62nd Annual Meeting of the Association
  for Computational Linguistics (Volume 1: Long Papers)}, pages 9796--9810,
  2024.

\bibitem[Chen et~al.(2025)Chen, Hu, Bai, Luo, Xue, Ren, Bai, Xie, Chen, Liu,
  et~al.]{astranav}
Jintao Chen, Junjun Hu, Haochen Bai, Minghua Luo, Xinda Xue, Botao Ren, Chengyu
  Bai, Shichao Xie, Ziyi Chen, Fei Liu, et~al.
\newblock Astranav-world: World model for foresight control and consistency.
\newblock \emph{arXiv preprint arXiv:2512.21714}, 2025.

\bibitem[Chen et~al.(2021)Chen, Guhur, Schmid, and Laptev]{history}
Shizhe Chen, Pierre-Louis Guhur, Cordelia Schmid, and Ivan Laptev.
\newblock History aware multimodal transformer for vision-and-language
  navigation.
\newblock \emph{NeurIPS}, 34:\penalty0 5834--5847, 2021.

\bibitem[Chen et~al.(2022)Chen, Guhur, Tapaswi, Schmid, and
  Laptev]{Thinkglobalactlocal}
Shizhe Chen, Pierre-Louis Guhur, Makarand Tapaswi, Cordelia Schmid, and Ivan
  Laptev.
\newblock Think global, act local: Dual-scale graph transformer for
  vision-and-language navigation.
\newblock In \emph{CVPR}, pages 16516--16526. IEEE, 2022.

\bibitem[Cheng et~al.(2024)Cheng, Ji, Yang, Gongye, Zou, Kautz, B{\i}y{\i}k,
  Yin, Liu, and Wang]{cheng2024navila}
An-Chieh Cheng, Yandong Ji, Zhaojing Yang, Zaitian Gongye, Xueyan Zou, Jan
  Kautz, Erdem B{\i}y{\i}k, Hongxu Yin, Sifei Liu, and Xiaolong Wang.
\newblock Navila: Legged robot vision-language-action model for navigation.
\newblock \emph{arXiv preprint arXiv:2412.04453}, 2024.

\bibitem[Chu et~al.(2026)Chu, Xie, Wu, Shen, Luo, Wang, Liu, Leng, Hu, Yin,
  et~al.]{abotn0}
Zedong Chu, Shichao Xie, Xiaolong Wu, Yanfen Shen, Minghua Luo, Zhengbo Wang,
  Fei Liu, Xiaoxu Leng, Junjun Hu, Mingyang Yin, et~al.
\newblock Abot-n0: Technical report on the vla foundation model for versatile
  embodied navigation.
\newblock \emph{arXiv preprint arXiv:2602.11598}, 2026.

\bibitem[Doshi et~al.(2024)Doshi, Walke, Mees, Dasari, and
  Levine]{scalingcrossembodied}
Ria Doshi, Homer Walke, Oier Mees, Sudeep Dasari, and Sergey Levine.
\newblock Scaling cross-embodied learning: One policy for manipulation,
  navigation, locomotion and aviation.
\newblock \emph{arXiv preprint arXiv:2408.11812}, 2024.

\bibitem[Gao et~al.(2026)Gao, Liang, Zheng, Malik, Ye, Yu, Tseng, Dong, Mo,
  Lin, et~al.]{gao2026dreamdojo}
Shenyuan Gao, William Liang, Kaiyuan Zheng, Ayaan Malik, Seonghyeon Ye, Sihyun
  Yu, Wei-Cheng Tseng, Yuzhu Dong, Kaichun Mo, Chen-Hsuan Lin, et~al.
\newblock Dreamdojo: A generalist robot world model from large-scale human
  videos.
\newblock \emph{arXiv preprint arXiv:2602.06949}, 2026.

\bibitem[Guo et~al.(2026)Guo, Wu, Duan, Ding, and Han]{guo2026phaselora}
Yufei Guo, Yinan Wu, Haoran Duan, Guiguang Ding, and Jungong Han.
\newblock Phaselora: Control-regime-conditioned low-rank adaptation for
  continuous-action vision-language-action policies.
\newblock \emph{arXiv preprint arXiv:2608.15285}, 2026.

\bibitem[Ha and Schmidhuber(2018)]{worldmodels}
David Ha and J{\"u}rgen Schmidhuber.
\newblock World models.
\newblock \emph{arXiv preprint arXiv:1803.10122}, 2\penalty0 (3):\penalty0 440,
  2018.

\bibitem[Hafner et~al.(2019{\natexlab{a}})Hafner, Lillicrap, Ba, and
  Norouzi]{dream}
Danijar Hafner, Timothy Lillicrap, Jimmy Ba, and Mohammad Norouzi.
\newblock Dream to control: Learning behaviors by latent imagination.
\newblock \emph{arXiv preprint arXiv:1912.01603}, 2019{\natexlab{a}}.

\bibitem[Hafner et~al.(2019{\natexlab{b}})Hafner, Lillicrap, Fischer, Villegas,
  Ha, Lee, and Davidson]{learning}
Danijar Hafner, Timothy Lillicrap, Ian Fischer, Ruben Villegas, David Ha,
  Honglak Lee, and James Davidson.
\newblock Learning latent dynamics for planning from pixels.
\newblock In \emph{ICML}, pages 2555--2565. PMLR, 2019{\natexlab{b}}.

\bibitem[Intelligence et~al.(2025)Intelligence, Black, Brown, Darpinian,
  Dhabalia, Driess, Esmail, Equi, Finn, Fusai, et~al.]{intelligence2025pi}
Physical Intelligence, Kevin Black, Noah Brown, James Darpinian, Karan
  Dhabalia, Danny Driess, Adnan Esmail, Michael Equi, Chelsea Finn, Niccolo
  Fusai, et~al.
\newblock $pi_{0.5}$: a vision-language-action model with open-world
  generalization.
\newblock \emph{arXiv preprint arXiv:2504.16054}, 2025.

\bibitem[Jeong and Yoon(2024)]{jeong2024quantifying}
Wooseong Jeong and Kuk-Jin Yoon.
\newblock Quantifying task priority for multi-task optimization.
\newblock In \emph{CVPR}, pages 363--372, 2024.

\bibitem[Jeong and Yoon(2025)]{jeong2025selective}
Wooseong Jeong and Kuk-Jin Yoon.
\newblock Selective task group updates for multi-task optimization.
\newblock In \emph{ICLR}, 2025.

\bibitem[Krantz et~al.(2020)Krantz, Wijmans, Majumdar, Batra, and
  Lee]{Beyondthenavgraph}
Jacob Krantz, Erik Wijmans, Arjun Majumdar, Dhruv Batra, and Stefan Lee.
\newblock Beyond the nav-graph: Vision-and-language navigation in continuous
  environments.
\newblock In \emph{ECCV}, pages 104--120. Springer, 2020.

\bibitem[Ku et~al.(2020)Ku, Anderson, Patel, Ie, and Baldridge]{rxr}
Alexander Ku, Peter Anderson, Roma Patel, Eugene Ie, and Jason Baldridge.
\newblock Room-across-room: Multilingual vision-and-language navigation with
  dense spatiotemporal grounding.
\newblock In \emph{Proceedings of the 2020 Conference on Empirical Methods in
  Natural Language Processing (EMNLP)}, pages 4392--4412, 2020.

\bibitem[Kullback and Leibler(1951)]{kullback1951information}
Solomon Kullback and Richard~A. Leibler.
\newblock On information and sufficiency.
\newblock \emph{The Annals of Mathematical Statistics}, 22\penalty0
  (1):\penalty0 79--86, 1951.

\bibitem[Lee and Zhao(2025)]{opentrackvla2025}
Kyusong Lee and Tiancheng Zhao.
\newblock Opentrackvla: Open-source visual language action model for visual
  navigation and following.
\newblock https://github.com/om-ai-lab/OpenTrackVLA, 2025.

\bibitem[Liang et~al.(2025)Liang, Czempin, Hong, Zhou, Wang, Biyik, and
  Tu]{clam}
Anthony Liang, Pavel Czempin, Matthew~M Hong, Yutai Zhou, Jingzhen Wang, Erdem
  Biyik, and Stephen Tu.
\newblock Clam: Continuous latent action models for robot learning from
  unlabeled demonstrations.
\newblock \emph{arXiv preprint arXiv:2505.04999}, 2025.

\bibitem[Lin et~al.(2025)Lin, Li, Zhao, Zhou, Wang, Wei, Tang, Li, Wang, Pang,
  et~al.]{vlnverse}
Sihao Lin, Zerui Li, Xunyi Zhao, Gengze Zhou, Liuyi Wang, Rong Wei, Rui Tang,
  Juncheng Li, Hanqing Wang, Jiangmiao Pang, et~al.
\newblock Vlnverse: A benchmark for vision-language navigation with versatile,
  embodied, realistic simulation and evaluation.
\newblock \emph{arXiv preprint arXiv:2512.19021}, 2025.

\bibitem[Liu et~al.(2025)Liu, Qi, Zhang, Li, Wang, Wu, Ye, Zhang, Chen, Zhong,
  et~al.]{trackvla++}
Jiahang Liu, Yunpeng Qi, Jiazhao Zhang, Minghan Li, Shaoan Wang, Kui Wu,
  Hanjing Ye, Hong Zhang, Zhibo Chen, Fangwei Zhong, et~al.
\newblock Trackvla++: Unleashing reasoning and memory capabilities in vla
  models for embodied visual tracking.
\newblock \emph{arXiv preprint arXiv:2510.07134}, 2025.

\bibitem[Liu et~al.(2024)Liu, Wang, and Yang]{volumetric}
Rui Liu, Wenguan Wang, and Yi Yang.
\newblock Volumetric environment representation for vision-language navigation.
\newblock In \emph{CVPR}, pages 16317--16328. IEEE, 2024.

\bibitem[Lu et~al.(2025)Lu, Zhou, Cheng, Meng, and Zheng]{Lu2025monoVLN}
Renjie Lu, Yu Zhou, Hao Cheng, Jingke Meng, and Wei-Shi Zheng.
\newblock monovln: Bridging the observation gap between monocular and panoramic
  vision and language navigation.
\newblock In \emph{ICCV}, pages 9477--9486, 2025.

\bibitem[Luo et~al.(2018)Luo, Sun, Zhong, Liu, Zhang, and Wang]{end}
Wenhan Luo, Peng Sun, Fangwei Zhong, Wei Liu, Tong Zhang, and Yizhou Wang.
\newblock End-to-end active object tracking via reinforcement learning.
\newblock In \emph{ICML}, pages 3286--3295. PMLR, 2018.

\bibitem[O’Neill et~al.(2024)O’Neill, Rehman, Maddukuri, Gupta, Padalkar,
  Lee, Pooley, Gupta, Mandlekar, Jain, et~al.]{openx}
Abby O’Neill, Abdul Rehman, Abhiram Maddukuri, Abhishek Gupta, Abhishek
  Padalkar, Abraham Lee, Acorn Pooley, Agrim Gupta, Ajay Mandlekar, Ajinkya
  Jain, et~al.
\newblock Open x-embodiment: Robotic learning datasets and rt-x models: Open
  x-embodiment collaboration 0.
\newblock In \emph{2024 IEEE International Conference on Robotics and
  Automation (ICRA)}, pages 6892--6903. IEEE, 2024.

\bibitem[Qiu et~al.(2017)Qiu, Zhong, Zhang, Qiao, Xiao, Kim, and
  Wang]{unrealcv}
Weichao Qiu, Fangwei Zhong, Yi Zhang, Siyuan Qiao, Zihao Xiao, Tae~Soo Kim, and
  Yizhou Wang.
\newblock Unrealcv: Virtual worlds for computer vision.
\newblock In \emph{ACM MM}, pages 1221--1224, 2017.

\bibitem[Team(2025)]{internvlan1}
IN Team.
\newblock Internvla-n1: An open dual-system vision-language navigation
  foundation model with learned latent plans.
\newblock Technical report, Technical report, 2025.

\bibitem[Team et~al.(2024)Team, Ghosh, Walke, Pertsch, Black, Mees, Dasari,
  Hejna, Kreiman, Xu, et~al.]{octo}
Octo~Model Team, Dibya Ghosh, Homer Walke, Karl Pertsch, Kevin Black, Oier
  Mees, Sudeep Dasari, Joey Hejna, Tobias Kreiman, Charles Xu, et~al.
\newblock Octo: An open-source generalist robot policy.
\newblock \emph{arXiv preprint arXiv:2405.12213}, 2024.

\bibitem[Wang et~al.(2025{\natexlab{a}})Wang, Xia, Zhao, Wang, Wang, Chen, Liu,
  Chen, and Pang]{vlnpe}
Liuyi Wang, Xinyuan Xia, Hui Zhao, Hanqing Wang, Tai Wang, Yilun Chen, Chengju
  Liu, Qijun Chen, and Jiangmiao Pang.
\newblock Rethinking the embodied gap in vision-and-language navigation: A
  holistic study of physical and visual disparities.
\newblock In \emph{2025 IEEE/CVF International Conference on Computer Vision
  (ICCV)}, pages 9455--9465. IEEE, 2025{\natexlab{a}}.

\bibitem[Wang et~al.(2025{\natexlab{b}})Wang, Zhang, Li, Liu, Li, Wu, Zhong,
  Yu, Zhang, and Wang]{trackvla}
Shaoan Wang, Jiazhao Zhang, Minghan Li, Jiahang Liu, Anqi Li, Kui Wu, Fangwei
  Zhong, Junzhi Yu, Zhizheng Zhang, and He Wang.
\newblock Trackvla: Embodied visual tracking in the wild.
\newblock \emph{arXiv preprint arXiv:2505.23189}, 2025{\natexlab{b}}.

\bibitem[Wang et~al.(2023)Wang, Li, Yang, Liu, and Jiang]{Gridmm}
Zihan Wang, Xiangyang Li, Jiahao Yang, Yeqi Liu, and Shuqiang Jiang.
\newblock Gridmm: Grid memory map for vision-and-language navigation.
\newblock In \emph{ICCV}, pages 15579--15590. IEEE, 2023.

\bibitem[Wei et~al.(2025)Wei, Wan, Yu, Wang, Yang, Mao, Zhu, Cai, Wang, Chen,
  et~al.]{wei2025streamvln}
Meng Wei, Chenyang Wan, Xiqian Yu, Tai Wang, Yuqiang Yang, Xiaohan Mao,
  Chenming Zhu, Wenzhe Cai, Hanqing Wang, Yilun Chen, et~al.
\newblock Streamvln: Streaming vision-and-language navigation via slowfast
  context modeling.
\newblock \emph{arXiv preprint arXiv:2507.05240}, 2025.

\bibitem[Xie et~al.(2026)Xie, Zhou, Zuo, Lu, Huang, Ma, and Yang]{uss}
Yuchen Xie, Xinyu Zhou, Kuangji Zuo, Yanshuo Lu, Fengrui Huang, Boyu Ma, and
  Jianfei Yang.
\newblock Uss: Unified spatial-semantic prompts for embodied visual tracking
  with latent dynamics learning.
\newblock \emph{arXiv preprint arXiv:2606.25880}, 2026.

\bibitem[Yao et~al.(2025)Yao, Gao, and Xu]{navmorph}
Xuan Yao, Junyu Gao, and Changsheng Xu.
\newblock Navmorph: A self-evolving world model for vision-and-language
  navigation in continuous environments.
\newblock In \emph{ICCV}, pages 5536--5546. IEEE, 2025.

\bibitem[Yao et~al.(2026)Yao, Zhu, Gao, Wang, and Xu]{yao2026sc}
Xuan Yao, Yuze Zhu, Junyu Gao, Zongmeng Wang, and Changsheng Xu.
\newblock Sc$^2$-wm: A self-correcting world model with closed-loop feedback
  for vision-and-language navigation in continuous environments.
\newblock \emph{arXiv preprint arXiv:2608.07548}, 2026.

\bibitem[Ye et~al.(2025)Ye, Jang, Jeon, Joo, Yang, Peng, Mandlekar, Tan, Chao,
  Lin, et~al.]{latent}
Seonghyeon Ye, Joel Jang, Byeongguk Jeon, Se~June Joo, Jianwei Yang, Baolin
  Peng, Ajay Mandlekar, Reuben Tan, Yu-Wei Chao, Bill~Yuchen Lin, et~al.
\newblock Latent action pretraining from videos.
\newblock In \emph{ICLR}, pages 28213--28239, 2025.

\bibitem[Zhang et~al.(2026{\natexlab{a}})Zhang, Zhang, Feng, Xiao, Sun, Chen,
  and Wang]{zhang2026mixture}
Boxuan Zhang, Weipu Zhang, Zhaohan Feng, Wei Xiao, Jian Sun, Jie Chen, and Gang
  Wang.
\newblock Mixture-of-world models: Scaling multi-task reinforcement learning
  with modular latent dynamics.
\newblock In \emph{ICLR}, 2026{\natexlab{a}}.

\bibitem[Zhang et~al.(2025)Zhang, Wang, Wang, Li, Liu, Wei, Wang, Zhang, and
  Wang]{zhang2024uni}
Jiazhao Zhang, Kunyu Wang, Shaoan Wang, Minghan Li, Haoran Liu, Songlin Wei,
  Zhongyuan Wang, Zhizheng Zhang, and He Wang.
\newblock Uni-navid: A video-based vision-language-action model for unifying
  embodied navigation tasks.
\newblock In \emph{Robotics: Science and Systems}, 2025.

\bibitem[Zhang et~al.(2026{\natexlab{b}})Zhang, Li, Qi, Li, Liu, Wang, Liu,
  Zhou, Wu, Li, et~al.]{embodied}
Jiazhao Zhang, Anqi Li, Yunpeng Qi, Minghan Li, Jiahang Liu, Shaoan Wang,
  Haoran Liu, Gengze Zhou, Yuze Wu, Xingxing Li, et~al.
\newblock Embodied navigation foundation model.
\newblock In \emph{ICLR}, pages 127293--127322, 2026{\natexlab{b}}.

\bibitem[Zhang et~al.(2026{\natexlab{c}})Zhang, Li, Qi, Li, Liu, Wang, Liu,
  Zhou, Wu, Li, et~al.]{zhang2026embodied}
Jiazhao Zhang, Anqi Li, Yunpeng Qi, Minghan Li, Jiahang Liu, Shaoan Wang,
  Haoran Liu, Gengze Zhou, Yuze Wu, Xingxing Li, et~al.
\newblock Embodied navigation foundation model.
\newblock In \emph{ICLR}, pages 127293--127322, 2026{\natexlab{c}}.

\bibitem[Zhang et~al.(2026{\natexlab{d}})Zhang, Zhou, Yin, Huang, Lei, Peng,
  Yuan, Zhang, Guo, Chen, et~al.]{qwenrobotnav}
Jiazhao Zhang, Gengze Zhou, Hale Yin, Yiyang Huang, Zixing Lei, Qihang Peng,
  Haoqi Yuan, Jie Zhang, Xudong Guo, Xiaoyue Chen, et~al.
\newblock Qwen-robotnav technical report: A scalable navigation model designed
  for an agentic navigation system.
\newblock \emph{arXiv preprint arXiv:2606.18112}, 2026{\natexlab{d}}.

\bibitem[Zhang et~al.(2026{\natexlab{e}})Zhang, Gong, Hao, Fu, Zhang, Zhou, Ye,
  Liang, Liang, and Ding]{zhang2026futurenav}
Lingfeng Zhang, Zeying Gong, Xiaoshuai Hao, Haoxiang Fu, Qiang Zhang, Mingliang
  Zhou, Hangjun Ye, Xiaojun Liang, Junwei Liang, and Wenbo Ding.
\newblock Futurenav: Unified world-action modeling for vision-and-language
  navigation.
\newblock \emph{arXiv preprint arXiv:2606.30367}, 2026{\natexlab{e}}.

\bibitem[Zhong et~al.(2019{\natexlab{a}})Zhong, Sun, Luo, Yan, and
  Wang]{zhong2019ad}
Fangwei Zhong, Peng Sun, Wenhan Luo, Tingyun Yan, and Yizhou Wang.
\newblock Ad-vat: An asymmetric dueling mechanism for learning visual active
  tracking.
\newblock In \emph{ICLR}, 2019{\natexlab{a}}.

\bibitem[Zhong et~al.(2019{\natexlab{b}})Zhong, Sun, Luo, Yan, and
  Wang]{zhong2019ad+}
Fangwei Zhong, Peng Sun, Wenhan Luo, Tingyun Yan, and Yizhou Wang.
\newblock Ad-vat+: An asymmetric dueling mechanism for learning and
  understanding visual active tracking.
\newblock \emph{IEEE TPAMI}, 43\penalty0 (5):\penalty0 1467--1482,
  2019{\natexlab{b}}.

\bibitem[Zhong et~al.(2024)Zhong, Wu, Ci, Wang, and Chen]{zhong2024EVT}
Fangwei Zhong, Kui Wu, Hai Ci, Churan Wang, and Hao Chen.
\newblock Empowering embodied visual tracking with visual foundation models and
  offline rl.
\newblock In \emph{ECCV}, pages 139--155. Springer, 2024.

\bibitem[Zhong et~al.(2025)Zhong, Wu, Wang, Chen, Ci, Li, and Wang]{unrealzoo}
Fangwei Zhong, Kui Wu, Churan Wang, Hao Chen, Hai Ci, Zhoujun Li, and Yizhou
  Wang.
\newblock Unrealzoo: Enriching photo-realistic virtual worlds for embodied ai.
\newblock In \emph{2025 IEEE/CVF International Conference on Computer Vision
  (ICCV)}, pages 5769--5779. IEEE, 2025.

\bibitem[Zhou et~al.(2024{\natexlab{a}})Zhou, Hong, Wang, Wang, and
  Wu]{navgpt2}
Gengze Zhou, Yicong Hong, Zun Wang, Xin~Eric Wang, and Qi Wu.
\newblock Navgpt-2: Unleashing navigational reasoning capability for large
  vision-language models.
\newblock In \emph{ECCV}, pages 260--278. Springer, 2024{\natexlab{a}}.

\bibitem[Zhou et~al.(2024{\natexlab{b}})Zhou, Hong, and Wu]{navgpt}
Gengze Zhou, Yicong Hong, and Qi Wu.
\newblock Navgpt: Explicit reasoning in vision-and-language navigation with
  large language models.
\newblock In \emph{AAAI}, pages 7641--7649, 2024{\natexlab{b}}.

\end{thebibliography}
}


\end{document}